\documentclass[journal]{IEEEtran} 
\IEEEoverridecommandlockouts
\usepackage{changes}
\usepackage{cite}
\usepackage{amsmath,amssymb,amsfonts}
\usepackage{algorithmic}
\usepackage{graphicx}
\usepackage{textcomp}
\usepackage{wrapfig}
\usepackage{bm}
\usepackage{algorithm}
\usepackage{multirow}
\usepackage{soul}
\usepackage{hyperref}
\usepackage{pdflscape}
\usepackage{multirow}

\usepackage{xcolor}

\title{\LARGE \bf
SMART-TRACK: A Novel Kalman Filter-Guided Sensor Fusion For Robust UAV Object Tracking in Dynamic Environments}

\author{\IEEEauthorblockN{Khaled Gabr\IEEEauthorrefmark{1},
Mohamed Abdelkader\IEEEauthorrefmark{1},\IEEEauthorrefmark{2},
Imen Jarraya\IEEEauthorrefmark{1}, 
Abdullah AlMusalami\IEEEauthorrefmark{1},
Anis Koubaa\IEEEauthorrefmark{1}}

\IEEEauthorblockA{\IEEEauthorrefmark{1} College of Computer and Information Sciences.
        Prince Sultan University. Riyadh, Saudi Arabia}\\
        \IEEEauthorrefmark{2} Corresponding author}

\newcommand{\keywords}[1]{\textbf{\textit{Keywords---}} #1}

\begin{document}

\maketitle
\thispagestyle{empty}
\pagestyle{empty}



\begin{abstract}
In the field of sensor fusion and state estimation for object detection and localization, ensuring accurate tracking in dynamic environments poses significant challenges. Traditional methods like the Kalman Filter (KF) often fail when measurements are intermittent, leading to rapid divergence in state estimations. To address this, we introduce SMART (Sensor Measurement Augmentation and Reacquisition Tracker), a novel approach that leverages high-frequency state estimates from the KF to guide the search for new measurements, maintaining tracking continuity even when direct measurements falter. This is crucial for dynamic environments where traditional methods struggle. Our contributions include: 1) Versatile Measurement Augmentation Using KF Feedback: We implement a versatile measurement augmentation system that serves as a backup when primary object detectors fail intermittently. This system is adaptable to various sensors, demonstrated using depth cameras where KF's 3D predictions are projected into 2D depth image coordinates, integrating nonlinear covariance propagation techniques simplified to first-order approximations. 2) Open-source ROS2 Implementation: We provide an open-source ROS2 implementation of the SMART-TRACK framework, validated in a realistic simulation environment using Gazebo and ROS2, fostering broader adaptation and further research. Our results showcase significant enhancements in tracking stability, with estimation RMSE as low as 0.04 m during measurement disruptions, advancing the robustness of UAV tracking and expanding the potential for reliable autonomous UAV operations in complex scenarios. The implementation is available at \url{https://github.com/mzahana/SMART-TRACK}.
\end{abstract}



\section{Introduction}\label{sec:intro}
\IEEEPARstart{U}{nmanned} Aerial Vehicles (UAVs) have become pivotal across diverse sectors, including aerial surveillance, urgent cargo delivery, and notably in disaster management \cite{kummritz2024sound, mohsan2023unmanned, mohsan2022towards}. This paper explores the development of an advanced state estimation framework for the detection and tracking of moving objects in three dimensions, a capability that is crucial across these applications \cite{nie20233d, karle2023multi, telli2023comprehensive}. The accurate and efficient tracking and state estimation of agile targets, such as UAVs, presents a significant challenge. The challenge is compounded by unpredictable environmental changes and intermittent noisy measurements from sensors like depth cameras or LiDARs at high speeds \cite{laghari2024unmanned, pliska2024towards, vrba2019onboard}. These issues can introduce noise and discontinuities, compromising the continuous and robust detection by AI-based object detectors and the effectiveness of traditional tracking systems\cite{tang2023n}.

Object detection, a fundamental task in the field of computer vision, has undergone a remarkable transformation, evolving from traditional methodologies to the forefront of deep learning techniques that currently define the discipline \cite{liu2020deep, zhao2019object}. Historically, object detection relied on multi-scale sliding window techniques, which involved systematically moving windows of various dimensions across an image while extracting manually engineered features \cite{tian2019fcos}. Classification algorithms such as Support Vector Machines (SVM) or AdaBoost were then employed to categorize the extracted features \cite{zou2019object}. However, these techniques were limited in robustness, accuracy, and computational complexity, prompting the demand for more sophisticated solutions \cite{liu2020deep}.

The advent of deep learning marked a pivotal moment in object detection. Convolutional neural networks (CNNs) emerged as a transformative force, allowing the learning of distinctive features directly from raw image data, eliminating the need for labor-intensive manual feature engineering \cite{zhang2020bridging}. Notable deep learning-based paradigms include the Single Shot Detector (SSD) family, exemplified by the influential You Only Look Once (YOLO) series \cite{redmon2019yolov3}. Variants such as YOLOv2, YOLOv3, YOLOv4, and YOLOv5 treat object detection as a regression problem, predicting both bounding box coordinates and class probabilities directly from the entire image in a single evaluation \cite{bochkovskiy2020yolov4}. This real-time performance and straightforward implementation have made these methods prominent, though they exhibit variations in accuracy across different versions \cite{wang2021scaled}.

Simultaneously, Two-Stage Detectors such as Faster R-CNN have become significant within deep learning-based object detection methods \cite{ren2019faster}. Faster R-CNN introduces a region proposal network (RPN) that generates region proposals while sharing convolutional features with the detection network. Although this two-stage structure improves precision, it increases computational complexity, necessitating a balance between accuracy and computational demands \cite{zhang2020bridging}.

The integration of various detection modalities is essential for achieving robust UAV tracking, especially in complex environments. Different sensor technologies, such as air surveillance radar, drone detection radar, vision-based systems, and acoustic sensors, cover various operational ranges and address different detection challenges. Air surveillance radars offer broad coverage, making them suitable for long-range detection, while drone detection radars provide medium-range tracking, and vision-based systems excel in close-range identification tasks \cite{singh2020drone, mohanti2021survey}. Acoustic sensors, although typically used in short-range scenarios, add another layer of detection that can operate in environments where visual and radar methods may struggle, such as in low-visibility conditions \cite{mallari2019multi, park2021survey}. Moreover, the integration of multi-modal sensors is crucial not only for detecting UAVs but also for ensuring their neutralization when necessary, thereby enhancing the overall security and operational effectiveness of UAV systems \cite{sharma2019comprehensive}.

Despite recent advances in tracking technologies, several gaps remain open and hinder their efficacy in real-world scenarios. Current state-of-the-art object detection systems, which heavily rely on deep learning algorithms like YOLOv8, excel in real-time and high-precision tasks \cite{xue2024yolo}. However, they tend to falter when targets exhibit high mobility or when tracking conditions deteriorate rapidly, leading to delayed or inaccurate estimations \cite{karle2023multi}. Furthermore, the typical operational frequencies of these systems, though high, do not consistently match the dynamics of UAV movement, resulting in periodic inaccuracies and a lack of reliability in critical situations.

Addressing these challenges, this paper introduces SMART-TRACK, a dual-priority approach for UAV detection and tracking. The SMART-TRACK method focuses on real-time, precise object detection using  \textbf{a combination of object detection techniques with Kalman Filter estimators as a feedback for rapid measurement re-acquisition}. As situations demand, it seamlessly transitions to a sophisticated Kalman Filter (KF)-guided algorithm, specifically designed to handle the rapid dynamics typical of UAV movements. This adaptive method not only maintains high operational frequencies, around 100 Hz but also significantly enhances the accuracy of the estimations.

The paper argues that while traditional methodologies often struggle with latency and the unpredictability of aerial target paths—leading to suboptimal state estimations—the Kalman Filter emerges as a superior tool for precise state estimation. It is particularly effective due to its dynamic adaptability to the requirements of the tracking system \cite{hematulin2023trajectory, lizzio2023comparison, silva2024high}. Nonetheless, deploying the KF effectively in scenarios involving highly agile targets, such as UAVs, demands an innovative approach that can keep pace with the rapid dynamics of UAV operations.

To overcome the limitations posed by inconsistent measurement frequencies from conventional AI vision-based detection systems, our approach enhances the detection system's capability through the projection of high-frequency predictions generated by the KF. This proactive strategy uses these predictions to guide the detection system, ensuring the acquisition of rapid, reliable measurements, even when the target exhibits unpredictable or highly dynamic behavior. This integration guarantees that the detection system remains synchronous with the swift operational requirements of UAV tracking, thereby maintaining the integrity and accuracy of the state estimation process.

This paper's contribution is a robust framework that not only addresses the existing gaps in UAV detection and tracking but also introduces an innovative use of the Kalman Filter to enhance the performance and reliability of UAV tracking systems in dynamic and challenging environments. This dual-priority approach, blending cutting-edge machine learning with sophisticated filtering techniques, represents a significant advancement in the field of UAV state estimation.

\subsection{Main Contributions}
This paper addresses critical challenges in UAV object tracking, particularly in maintaining tracking continuity in complex, dynamic environments where sensor measurements may be intermittent or unreliable. Our key contributions are as follows:

\begin{itemize}
\item \textbf{Measurement Augmentation to Handle Sensor Failures}: Traditional object tracking systems often struggle when primary object detectors, such as YOLOv8, fail intermittently due to sensor limitations or environmental factors. To address this, we introduce a measurement augmentation system that leverages Kalman Filter (KF) predictions to generate search regions for acquiring new measurements. This approach is adaptable to various sensor types, and we demonstrate its effectiveness with depth cameras, where the KF's 3D mean and covariance predictions are projected into 2D depth image coordinates. Our method involves nonlinear covariance propagation techniques, simplified to first-order approximations, to ensure efficient and reliable measurement integration. This solution represents a significant advancement in maintaining tracking stability and precision despite sensor interruptions.

\item \textbf{Scalable and Open-source ROS2 Implementation}: Recognizing the need for accessible and scalable solutions, we have developed the SMART-TRACK framework, which has been validated in a realistic simulation environment using the Gazebo simulator and ROS2. This implementation not only facilitates seamless integration with existing ROS2-based systems but also promotes further research and development by being open-source. The framework is designed to be easily adaptable to other platforms and sensor types, ensuring broad applicability across various UAV tracking scenarios. The implementation can be viewed in action in a demonstration video available at \url{https://youtu.be/MtPVoIZme6k?feature=shared} and accessed for public use at \url{https://github.com/mzahana/SMART-TRACK}.
\end{itemize}

\section{Related Works} \label{sec:related_works}
The landscape of UAV detection and tracking is characterized by rapid advancements and the integration of complex technologies. This section critically analyzes key methodologies and technologies currently in use and closely related to our work, notably depth cameras and LiDAR, and how AI-based methods are shaping the evolution of UAV tracking systems \cite{nawaz2023robust, cao2023real}.

The evolution of UAV object detection and tracking has been remarkable, shaped by advances in both hardware and software technologies. In the early stages of computer vision during the 1960s and 1970s, research primarily focused on edge detection and simple geometric shapes \cite{rosenfeld1976digital, bennamoun2012object}. These foundational techniques laid the groundwork for more complex methods but were often insufficient for the dynamic and fast-paced nature of UAV operations.

The 1990s marked a significant milestone with the introduction of statistical models such as the Kalman Filter, which provided a framework for more accurate predictions of object movements in dynamic environments \cite{zaidi2022survey}. The Kalman Filter's ability to estimate the state of a system from noisy observations made it particularly beneficial for UAV applications, where precise state estimation is critical. This period also saw the emergence of other filtering techniques, such as the Extended Kalman Filter (EKF) and the Particle Filter, which further enhanced the robustness of tracking systems in complex environments.

The past decade has witnessed a revolutionary transformation in object detection and tracking, driven by the rise of deep learning. Convolutional Neural Networks (CNNs) have become the cornerstone of modern computer vision, leading to significant breakthroughs such as the You Only Look Once (YOLO) framework and Faster R-CNN \cite{vijayakumar2024yolo, xue2024yolo }. Recent advancements, including YOLOv5 and YOLOv8, have further refined these techniques, achieving remarkable improvements in real-time object detection and tracking \cite{sang2024environmentally }. These deep learning models leverage hierarchical feature extraction and end-to-end training, dramatically enhancing both speed and accuracy in complex scenarios.

However, despite these advancements, several gaps remain. Current state-of-the-art object detection systems, while excelling in real-time and high-precision tasks, often struggle with high mobility targets or deteriorating tracking conditions \cite{karle2023multi}. These systems' operational frequencies, although high, may not consistently align with the rapid dynamics of UAV movement, leading to periodic inaccuracies and reduced reliability in critical situations. The following sub-sections highlight the main literature and technologies closely related to our work including depth-based, LiDAR-based, and deep learning-based object detection.

\subsection{Depth-Based Drone Detection}
Depth cameras, employing technologies such as stereo vision, are pivotal in enhancing UAV detection capabilities by providing three-dimensional spatial information essential for accurate tracking \cite{hwang20233d}. Studies like \cite{8756100} illustrate the practical application of depth imagery in intercepting non-cooperative drones efficiently, even on platforms with limited computational resources. Despite their advantages, the performance of depth-based systems is often curtailed in low visibility and fluctuating lighting conditions, leading to inaccuracies and increased false positives \cite{wang2023adaptive, song2023synthetic}. These limitations underscore the necessity for more robust algorithms that can maintain performance integrity across a broader range of environmental conditions, prompting a reevaluation of traditional depth-sensing methodologies.

\subsection{LiDAR-Based Drone Detection}
LiDAR technology has been integral in pushing the boundaries of precision in UAV tracking due to its ability to generate detailed 3D environmental maps \cite{aldao2022lidar}. Innovations such as integrating LiDAR with 4D radar and imaging data have been pivotal in enhancing detection accuracy, as demonstrated by the "sampling" view transformation strategy in \cite{xiong2023lxl}. This approach improves the synergy between different sensor modalities, enhancing the robustness and reliability of UAV tracking systems. However, LiDAR's susceptibility to environmental interferences like adverse weather conditions poses significant challenges, often affecting the reliability of the data and, consequently, the tracking performance \cite{gomez2023efficient, lee2023design}. These challenges highlight the ongoing need for adaptive systems that can dynamically adjust to changing environmental parameters.

\subsection{AI-Based Methods}
Artificial Intelligence has revolutionized UAV detection with the advent of deep learning technologies like CNNs, RNNs, and advanced algorithms such as DRL and GANs \cite{zhang2023bionic, koubaa2023aero, bai2024improved, vrba2020marker}. These methods offer remarkable adaptability and have proven effective across varying operational conditions, significantly enhancing real-time tracking capabilities. The utilization of AI extends beyond mere detection to include autonomous decision-making and the generation of realistic synthetic training data, broadening the scope of UAV tracking applications. However, the dependency on high computational resources and extensive training datasets presents substantial hurdles, necessitating continuous advancements in computational efficiency and data management strategies \cite{sai2023comprehensive}.

\subsection{Our Combined Approach}
The SMART-TRACK framework represents a synthesis of the insights gained from evaluating depth-based, LiDAR-based, and AI-enhanced methodologies. By integrating state estimation feedback to guide new measurement acquisition during detection failures, we address a critical gap in existing technologies, thereby significantly enhancing tracking accuracy and system reliability. 
We go beyond the traditional object detection and state estimation frameworks by leveraging the state estimation as feedback to guide the search for new measurements whenever the object detector pipeline fails to do so. We show that this approach significantly increases the overall state estimation accuracy by an order of magnitude through experiments in realistic simulation environments. This approach is validated through extensive simulations, demonstrating its effectiveness in improving state estimation accuracy by an order of magnitude in dynamic and challenging environments.

\section{System Architecture: Preliminaries and Overview}\label{sec:problem}

\begin{figure}
    \centering
    \includegraphics[width=\columnwidth]{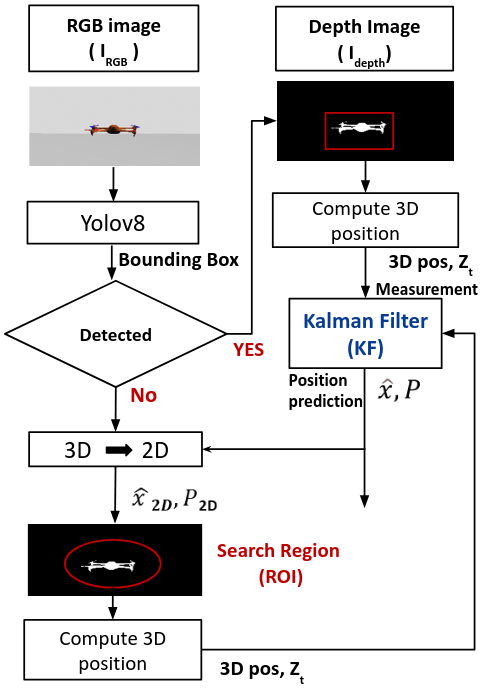}
    \caption{SMART-TRACK system flowchart}
    \label{fig:process_flowchart}
\end{figure}

In this section, we define the system preliminaries and provide an overview of the SMART-TRACK system components, and its details are described in section \ref{sec:algorithm}.

\subsection{Preliminaries}
\begin{figure}[ht]
    \centering
    \includegraphics[width=\columnwidth]{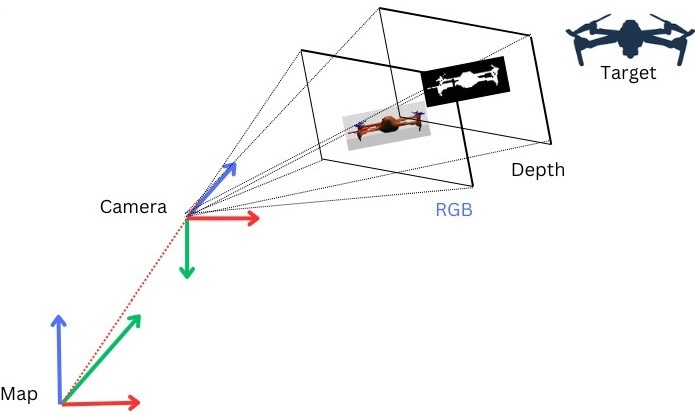}
    \caption{Target expressed in color (RGB) and depth image frames.  }
    \label{fig:frame}
\end{figure}

\begin{figure}[ht]
    \centering
\includegraphics[width=\columnwidth]{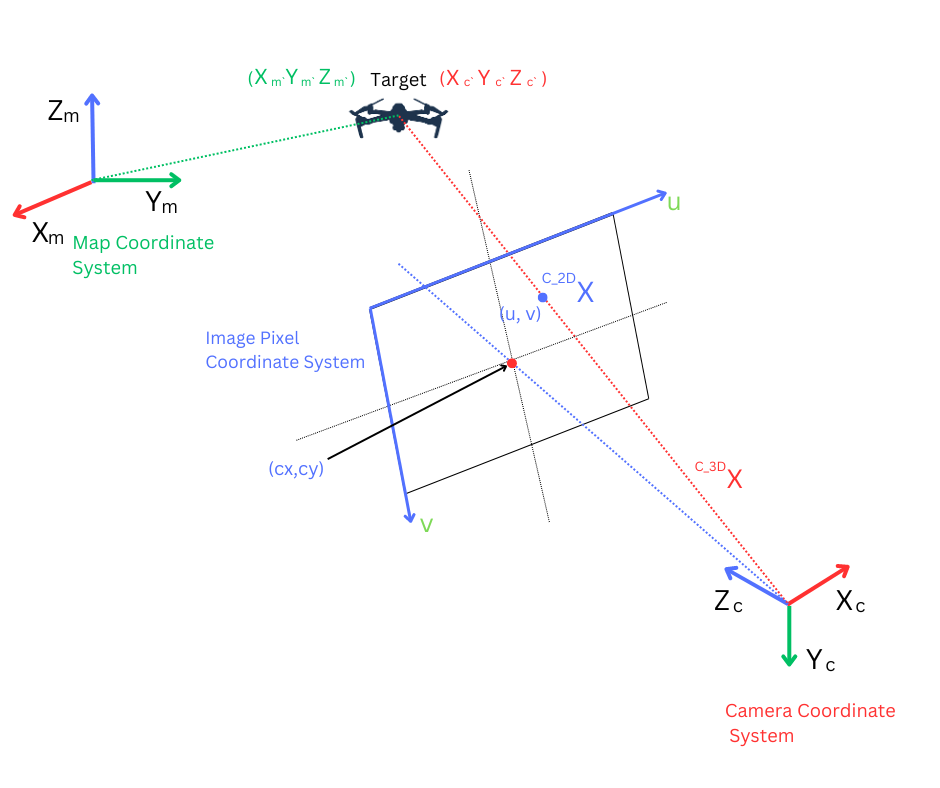}
    \caption{The target's position  expressed in map frame (green line), camera 3D coordinate frame (red line), and the camera 2D image frame ($^{C_{2D}}X$) }
    \label{fig:coordinate_frames}
\end{figure}

We address the challenge of estimating the 3D position and velocity of a moving target (i.e., an intruder UAV) at discrete time steps, denoted as $t$. The target state $\bm{x}_t$ comprises position and velocity in 3D, represented as $\bm{x}_t=[p_x, p_y, p_z, v_x, v_y, v_z]^T \in \mathbb{R}^6$. This state is considered within an inertial frame, termed the map frame, and is denoted as $^{M}\bm{x}_t$. Additionally, the target state can be expressed in the sensor's frame (e.g. camera), where detection occurs, denoted as $^{C}\bm{x}_t$. Fig. \ref{fig:frame} and \ref{fig:coordinate_frames} depict the different frames in which the target's position is expressed.

The target's state dynamics are system design choice. For simplicity, we assume the states follow a linear discrete-time model defined in equation \eqref{eq:dynamics}.
\begin{equation} \label{eq:dynamics}
^M\bm{x}_{t}=\bm{F} ^M\bm{x}_{t-1} + \bm{w}
\end{equation}

Here, $\bm{F}$ is the constant transition matrix, defined in relation to the system's prediction time $\Delta t$.
\begin{equation}\label{eq:F_matrix}
\bm{F} = \begin{bmatrix} 1 & 0 & 0 & \Delta t & 0 & 0 \\ 0 & 1 & 0 & 0 & \Delta t & 0 \\ 0 & 0 & 1 & 0 & 0 & \Delta t \\ 0 & 0 & 0 & 1 & 0 & 0 \\ 0 & 0 & 0 & 0 & 1 & 0 \\ 0 & 0 & 0 & 0 & 0 & 1 \end{bmatrix}
\end{equation}

The process noise $\bm{w}$ is typically modeled as Gaussian $\mathcal{N}(\bm{0}, \bm{Q})$, with a mean of $\bm{0}$ and covariance matrix $\bm{Q}$.

Using these definitions, the expected state and its covariance are predicted using the following standard Kalman Filter (KF) prediction equations.
\begin{equation}\label{eq:state_prepdiction}
^M\hat{\bm{x}}_{t|t-1} = \bm{F} ^M\hat{\bm{x}}_{t-1|t-1}
\end{equation}
\begin{equation}\label{eq:cov_prediction}
\bm{P}_{t|t-1} = \bm{F} \bm{P}_{t-1|t-1} \bm{F}^\top + \bm{Q}
\end{equation}

The measurement at each discrete time step $t$ is the observed position of the target in the map frame, denoted by $^M\bm{z}_t=[x_m, y_m, z_m]^T \in \mathbb{R}^3$. We focus on position observations reconstructed from depth measurements provided by a depth camera, employing a standard KF linear measurement model, equation \eqref{eq:measurement_model}.

\begin{equation}\label{eq:measurement_model}
^{M}\bm{z}_{t} = \bm{H} ^{M}\bm{x}_{t} + \bm{v}
\end{equation}
The measurement matrix $\bm{H}$ used is as follows:

\begin{equation}\label{eq:measurement_matrix}
\bm{H} = \begin{bmatrix} 1 & 0 & 0 & 0 & 0 & 0 \\ 0 & 1 & 0 & 0 & 0 & 0 \\ 0 & 0 & 1 & 0 & 0 & 0 \end{bmatrix}
\end{equation}

Here, the measurement noise $\bm{v}$ is assumed to be Gaussian, $\mathcal{N}(\bm{0}, \bm{R})$, with a mean of $\bm{0}$ and covariance matrix $\bm{R}$.



The KF state and covariance predictions are corrected using the measurements according to the following standard KF update equations.

\begin{equation}\label{eq:kalman_gain}
\bm{K}_t = \bm{P}_{t|t-1} \bm{H}^\top (\bm{H} \bm{P}_{t|t-1} \bm{H}^\top + \bm{R})^{-1}
\end{equation}

\begin{equation}\label{eq:state_update}
^{M}\hat{\bm{x}}_{t|t} = ^{M}\hat{\bm{x}}_{t|t-1} + \bm{K}_t (^{M}\bm{z}_t - \bm{H} ^{M}\hat{\bm{x}}_{t|t-1})
\end{equation}

\begin{equation}\label{eq:cov_update}
\bm{P}_{t|t} = (I - \bm{K}_t \bm{H}) \bm{P}_{t|t-1}
\end{equation}

where:
\begin{itemize}
    \item $\bm{K}_t$ is the Kalman gain at time $t$.
    \item $\bm{P}_{t|t-1}$ is the predicted state covariance matrix.
    \item $\bm{H}$ is the observation model matrix that maps the predicted state to the observed measurements.
    \item $\bm{R}$ is the measurement noise covariance matrix.
    \item $^{M}\hat{\bm{x}}_{t|t}$ is the updated state estimate at time $t$ after incorporating the measurement.
    \item $^{M}\hat{\bm{x}}_{t|t-1}$ is the predicted state estimate at time $t$ before incorporating the measurement.
    \item $^{M}\bm{z}_t$ is the measurement vector at time $t$.
    \item $I$ is the identity matrix.
\end{itemize}


These equations ensure that the state and covariance estimates are adjusted based on the incoming measurements, enhancing the accuracy and reliability of the tracking system.

In practice, the predictions of the expected state and corresponding covariance,  defined in \eqref{eq:state_prepdiction} and (\ref{eq:cov_prediction}), are utilized in subsequent control and tracking tasks rather than the corrected states. These predictions are computed at a higher frequency than the corrections and are updated as soon as a valid measurement is received. However, the quality of these predictions depends on the quality and frequency of previous measurements. Frequent loss of measurements can lead to divergence in the KF estimates. To counter this, our approach uses high-frequency KF measurements as feedback to expedite the search for new measurements when primary measurements from the object detection module are unavailable. 

Our current focus is on maintaining a balance between performance and computational feasibility, ensuring our approach remains effective and efficient in real-time applications. The standard KF was selected due to its simplicity, lower computational demands, and suitability for the linear dynamics typically encountered in high-speed target tracking. While the Extended Kalman Filter (EKF) and Unscented Kalman Filter (UKF) offer superior handling of non-linear dynamics, they require more complex mathematical operations and precise parameter tuning, which can significantly increase computational overhead. In our application, where real-time performance is critical, the trade-offs associated with EKF and UKF would likely outweigh their benefits, making the standard KF the more practical choice.

In this paper, our focus is on detecting and localizing a target UAV using depth and RGB cameras, although the concept is equally applicable with a combination of calibrated 3D LiDAR and RGB camera. The RGB image facilitates object detection, while the depth image (or 3D LiDAR data) aids in localizing the object within the bounding box provided by the object detection module. We operate under the assumption that the depth and RGB camera frames are synchronized and aligned, ensuring that both the RGB image frame, $I_{rgb}$, and the depth image frame, $I_{depth}$, correspond to the same time instant $t$. This synchronization is crucial for utilizing the bounding box consistently across both images, with alignment ensuring that $I_{depth}$ is expressed in the coordinate frame of $I_{rgb}$.

In scenarios where the object is visible in both $I_{rgb}$ and $I_{depth}$, but not detected in $I_{rgb}$ by the object detector, we utilize the Kalman filter's predictions. For a target $i$, the predictions $^M\hat{\bm{x}}_t$ and $\bm{P}_t$ are used to project a 2D search region, specifically an ellipse denoted by $E_i$, onto $I_{depth}$. A new measurement, labeled $^{KF}z_t$ (representing the target's 3D position), is then constructed using $E_i$, as detailed in section \ref{sec:algorithm}. This measurement is subsequently employed by the Kalman Filter to update its current predictions $^{M}\hat{\bm{x}}_t$ and $\bm{P}_t$.

The following subsection provide an overview of the SMART-TARCK framework and the details are discussed in section section \ref{sec:algorithm}.

\subsection{SMART-TRACK Framework Overview}
Traditional object tracking methods are composed of two main pipelines: 
\begin{itemize}
    \item \textit{\textbf{Object detection:}} This is performed using several techniques based on the sensor type such as LiDAR, RADAR, and vision sensors. Most common object detection techniques use state-of-the-art AI vision-based methods. These methods can be categorized into two categories \cite{zaidi2021survey}:  single-stage (such as YOLOv4 \cite{bochkovskiy2020yolov4}) and tow-stage (such as Mask R-CNN \cite{he2018mask}) detectors. In this work, we use YOLOv8 \cite{Jocher_YOLO_by_Ultralytics_2023}, a single-stage detector known for its real-time performance. The output of the object detector is then transformed into a measurement that is fed to the state estimation pipeline.
    
    \item \textit{\textbf{State estimation:}} This is the process of estimating the true state of the tracked object, such as position, velocity, and acceleration. This is commonly done using a Kalman filter framework, which predicts the probability distribution of the system states and its uncertainty using a model of choice. Then, the predictions are corrected using real measurements received from the object detector.
\end{itemize}

The SMART-TRACK framework extends the previous steps as follows.
\begin{itemize}
    \item \textit{\textbf{Search region proposal:}} Whenever the primary object detector fails to provide measurements, the Kalman filter estimates (mean and covariance) are projected onto the sensor coordinate frame to construct search regions (ellipses) in which the tracked target is expected to be.
    \item \textit{\textbf{Measurement re-acquisition:}} The search regions are then used to extract a new object measurement (position) that is fed to the Kalman filter to update its prediction.
\end{itemize}
This feedback mechanism is the core concept of the SMART-TRACK framework, which ensures measurement continuity, and, therefore, accurate and stable estimation.

This framework is agnostic to the type of sensor as long as there is a mechanism of transforming the KF estimates to the sensor frame. However, in this work we consider a depth camera sensor as the search region proposal (projection of 3D KF estimates onto 2D depth image) is not trivial. For a 3D LiDAR sensor the projection is trivial as it is 3D to 3D. Otherwise the same steps applies as described in section \ref{sec:algorithm}.

Using a depth camera sensor, the SMART-TRACK framework workflow is summarized as follows. The RGB camera input is processed using YOLOv8 to detect a target of interest, a drone in this work. The bounding box of the detection is used to find the corresponding depth pixels in the depth image. These pixels, in addition to the camera intrinsics, are used to reconstruct the object 3D location with respect to the camera. If this detection process succeeds, the 3D location is fed to the Kalman filter for state estimation. If it fails, due to environmental factors or model inaccuraciesm the latest KF estimates are projected onto the depth frame (3D to 2D projection) to propose search region. The search region is then used to extract new 3D measurements for the KF to consume to correct its predictions. This process is illustrated in Fig. \ref{fig:process_flowchart}.

This 'KF-guided measurement' compensates for gaps in object detection, maintaining the precision of the UAV tracking in dynamic environments. Thus, the Kalman filter not only offers high-frequency state estimates but also aids in object re-detection using depth camera data, forming a dual-functional core in our innovative 3D target tracking approach.

\section{SMART-TRACK: Sensor Measurement Augmentation and Reacquisition Tracker}\label{sec:algorithm}
This section details our SMART-TRACK frawework, designed to enhance Kalman filter state estimation for tracking a target's (in this case, a UAV) 3D trajectory. Our goal is to ensure a stable stream of measurements for the Kalman filter, enabling frequent correction of its predictions, even when vision-based target detection is intermittent. The overall process is outlined in Fig. \ref{fig:process_flowchart}, with further details in the subsequent subsections.

The target tracking system begins with the detection of the target object. Although we define the detection method based on a depth camera it equally applies on a LiDAR-based object detection and localization. The considered input measurement in this work is an RGB image denoted by $I_{rgb}$, where the target is detected and marked by a bounding box using a real-time object detection convolutional neural network, such as YOLOv8.

This bounding box from $I_{rgb}$ is then applied to the corresponding depth image denoted by $I_{depth}$ to estimate the target's 3D position relative to the camera frame ($^C\bm{z}_t$ at time step $t$). This estimate is transformed to the map frame ($^M\bm{z}_t = ^MT_C \cdot ^C\bm{z}_t$) and fed to the Kalman filter for prediction correction equations in \eqref{eq:state_update} and \ref{eq:cov_update}. This procedure is detailed in Algorithm \ref{alg:detection_to_pose}.

Upon successful prediction by the KF, if the target is visible (available in both RGB and depth images) but not detected (using the object detector), the KF-guided measurement search commences. Here, the latest KF prediction of the target's position and covariance ($^M\hat{\bm{x}}_t$ and $\bm{P}_t$) is used to construct a 2D search region (ellipse $E$) in $I_{depth}$, as outlined in Algorithm \ref{alg:search_region_extraction}. Within this region, we identify valid depth pixels representing the target to compute a new measurement of its 3D position ($^{KF}\bm{z}_{t}$), which is then used to update the Kalman filter predictions. This step is summarized in Algorithm \ref{alg:measurement_extraction}.

\subsection{Target Detection and Localization}

While the primary focus of this paper is not on UAV detection, we briefly address it for the sake of completeness in our approach to UAV tracking. Our method utilizes a custom YOLOv8 model \cite{Jocher_YOLO_by_Ultralytics_2023}, chosen for its effectiveness in real-time object detection. This model, trained on a diverse dataset of 41,729 images, including data from various sources \cite{drone-5eg4c_dataset, drone-detection-q5xwg_dataset} as well as our own environmental recordings, specializes in identifying multi-rotor UAVs.

The detected UAVs in the RGB image frame $I_{rgb}$ are marked with bounding boxes. These boxes correspond to groups of pixels within the depth image $I_{depth}$, which are then utilized to compute the 3D position of the target in the camera frame, $^C\bm{z}_t$ (equation \eqref{eq:yolo_3D_measurement_cam_frame}). This data is subsequently transformed to the map frame $M$ using  \ref{eq:yolo_3D_measurement_map_frame} and used to correct the predictions of the Kalman filter. It's important to note that this aspect of detection is a supporting component of our overarching target tracking methodology.

 \begin{equation} \label{eq:yolo_3D_measurement_cam_frame}
    ^C\bm{z}_t =
    \begin{bmatrix}
    X \\
    Y \\
    Z
    \end{bmatrix} = d(x_c, y_c) \cdot \bm{K}^{-1} \cdot \begin{bmatrix}
    x_c \\
    y_c \\
    1
    \end{bmatrix},
\end{equation}

\begin{equation}\label{eq:yolo_3D_measurement_map_frame}
 \quad ^{M}\bm{z}_t = ^{M}T_C \cdot ^C\bm{z}_t
\end{equation}

Where $d(x_c, y_c)$ is the depth value at the bounding box center in the $I_{rgb}$ frame, $\bm{K}$ is the camera intrinsic parameters matrix, and $^MT_C$ is the homogeneous transformation matrix from the camera frame to the map frame. Object localization using the detected object is summarized in Algorithm \ref{alg:detection_to_pose}.

\begin{algorithm}
\caption{Object Localization from RGB/Depth Images}
\label{alg:detection_to_pose}
\begin{algorithmic}
\STATE \textbf{Input}: RGB image $I_{rgb}$, Depth image $I_{depth}$ , Camera intrinsic parameters $\bm{K}$, Bounding Box $B$
\STATE \textbf{Output}: $^M\bm{z}_t$ target position measurement in the map frame
\STATE \textbf{do}:
\STATE - $B \leftarrow$ apply object detection on $I_{rgb}$ and return bound box
\STATE - $(x_c, y_c) \leftarrow$ Extract center of $B$
\STATE - $d(x_c,y_c) \leftarrow$ Calculate average depth within $B$
\STATE - $^C\bm{z}_t \leftarrow$ Calculate 3D pose in the camera frame using Equation \ref{eq:yolo_3D_measurement_cam_frame}
\STATE - $^M\bm{z}_t \leftarrow$ Transform position to map frame using \eqref{eq:yolo_3D_measurement_map_frame}
\STATE - return $^M\bm{z}_t$
\STATE \textbf{end}
\end{algorithmic}
\end{algorithm}

\subsection{Extracting New Measurements Using KF Estimates}
As shown in Fig. \ref{fig:process_flowchart}, whenever the primary target detection fails, the KF-guided measurement search process begins. This process starts by using the latest Kalman filter estimates to construct a 2D search region, an ellipse $E$, in $I_{depth}$. The search region $E$ is constructed using the 2D projection of the 3D KF estimates ($^M\hat{\bm{x}}_t, \bm{P}_t$) onto  $I_{depth}$, resulting in 2D estimates ($\hat{\bm{x}}_{2D}, \bm{P}_{2D}$). The new 3D position measurements $^{KF}\bm{z}$ are searched for and calculated  within the constructed search region $E$, and upon re-detection, these measurements are sent to the KF to update the UAV's 3D position in the map frame. This process exemplifies the robustness of our system in maintaining accurate UAV tracking even if there is discontinuity in the object detection module.

The mathematical foundation for projecting 3D points and covariance onto the 2D image plane is established as follows.

\subsubsection{Projection of a 3D KF mean onto Depth Image}

Consider the mean 3D position $^M\bm{\mu}$ estimated by the KF in the map frame. Using standard perspective geometry, this point can be transformed to the 3D camera frame using $^{C_{3D}}\bm{\mu}= [p_x, p_y, p_z]^T = ^{C_{3D}}T_M \cdot  ^M\bm{\mu}$. The projection of $^{C_{3D}}\bm{\mu}$  onto $I_{depth}$ yields the coordinates $ ^{C_{2D}}\bm{\mu} = [u, v]^T $ using the camera's intrinsic parameters, equation \eqref{eq:cam_projection}.

\begin{equation}\label{eq:cam_projection}
^{C_{2D}}\bm{\mu}=\\
\begin{bmatrix}
u \\
v
\end{bmatrix} = 
\begin{bmatrix}
f_x & 0 & c_x \\
0 & f_y & c_y \\
0 & 0 & 1
\end{bmatrix}
\begin{bmatrix}
\frac{p_x}{p_z} \\
\frac{p_y}{p_z} \\
1
\end{bmatrix}
\end{equation}

where $ f_x, f_y $ are the focal lengths along the image X and Y axes, and $ c_x, c_y $ are the coordinates of the principal point.\\

\subsubsection{Projection of 3D Covariance onto Depth Image}
The Kalman Filter's estimated 3D position covariance matrix in the map frame, $^M\bm{\Sigma}$, associated with $^{M}\bm{\mu}$, is transformed into the 3D camera frame as $^{C_{3D}}\bm{\Sigma} = {^{C_{3D}}T_M} \cdot {^M\bm{\Sigma}}$. This transformation is key to defining a 2D search region in $I_{depth}$, using the position's expected value and uncertainty computed by the Kalman Filter.

The conversion of 3D estimates ($^{C_{3D}}\bm{\mu}$, $^{C_{3D}}\bm{\Sigma}$) to 2D estimates in $I_{depth}$ coordinates is achieved through nonlinear covariance propagation, approximated to first order as per \cite{Ochoa2006CovariancePF}. The 2D projection of the 3D covariance matrix onto $I_{depth}$ is executed by computing the Jacobian matrix of the projection and implementing the transformation as specified in \eqref{eq:cov_3d_to_2d}.

\begin{equation}\label{eq:cov_3d_to_2d}
^{C_{2D}}\bm{\Sigma} = \bm{J} \cdot ^{C_{3D}}\bm{\Sigma} \cdot \bm{J}^T
\end{equation}

where the Jacobian matrix $ \bm{J} $ is derived according to \eqref{eq:jacobian}.

\begin{equation}\label{eq:jacobian}
\bm{J} = 
\begin{bmatrix}
\frac{\partial u}{\partial p_x} & \frac{\partial u}{\partial p_y} & \frac{\partial u}{\partial p_z} \\
\frac{\partial v}{\partial p_x} & \frac{\partial v}{\partial p_y} & \frac{\partial v}{\partial p_z}
\end{bmatrix} = 
\begin{bmatrix}
\frac{f_x}{p_z} & 0 & -\frac{f_x p_x}{p_z^2} \\
0 & \frac{f_y}{p_z} & -\frac{f_y p_y}{p_z^2}
\end{bmatrix}
\end{equation}

This formulation is integral to the algorithm's ability to adaptively define search regions for object re-detection, ensuring robust tracking performance even with intermittent measurements. The 2D estimation of ($^{C_{2D}}\bm{\mu}, ^{C_{2D}} \bm{\Sigma}$) resembles the KF estimates in terms of target's 3D position mean  and its covariance (uncertainty) in the 2D image $I_{depth}$, which is then used to construct the search ellipse $E$ as explained in the next section.

\subsubsection{Search Region Construction}

\begin{figure}[ht]
    \centering
    \includegraphics[width=0.7\columnwidth]{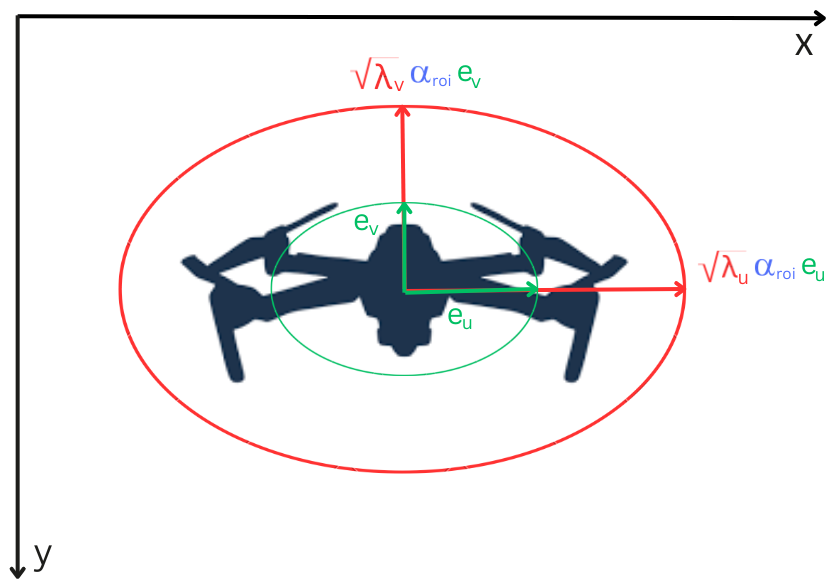}
    \caption{The search ellipse $E$ in the depth image $I_{depth}$. The original $E$ is constructed using the eigenvalues and eigenvectors of $^{C_{2D}}\bm{\Sigma}$ (green). The scaled ellipse $\bar{E}$ (red).}
    \label{fig:ROI}
\end{figure}

The search region (ellipse) $E$ is constructed using the eigenvalues and eigenvectors of the 2D projected covariance matrix $^{C_{2D}} \bm{\Sigma}$. The eigenvalues ($\lambda_u, \lambda_v$) represent the magnitude of uncertainty in each principal direction, while eigenvectors ($\bm{e}_u, \bm{e}_v$) indicate the orientation of these uncertainties  in the image plane. The eigenvectors are used to define the directions of the major and minor axes of the search ellipse $E$, while the eigenvalues can be used to define the size of $E$. The half-length of the major and minor axes of $E$ are $\sqrt{\lambda_u}$ and $\sqrt{\lambda_v}$, respectively, as seen in Fig. \ref{fig:ROI} . When the KF uncertainties are low (small $\lambda_u, \lambda_v$), the the size of $E$ can be arbitrarily scaled by $\alpha_{roi} > 0$, equation (\ref{eq:scaling}), to define a reasonable search region in which the target in the depth image $I_{depth}$ is expected to be seen. This scaler is a design parameter and depends on the target of interest.
\begin{equation} \label{eq:scaling}
    l_{u} = \alpha_{\text{roi}} \cdot \sqrt{\bm{\lambda_{u}}}, \quad l_v = \alpha_{\text{roi}} \cdot \sqrt{\bm{\lambda_{v}}}
\end{equation}

The construction of the search region $E$ is summarized in Algorithm \ref{alg:search_region_extraction} and depicted in Fig.\ref{fig:ROI} . 

\begin{algorithm}
\caption{Search Region Construction Using KF Feedback} \label{alg:search_region_extraction}
\begin{algorithmic}
\STATE \textbf{Input}: Kalman Filter position estimate  $^M \bm{\mu}$, 3D position covariance matrix $^M\bm{\Sigma}$, depth image $I_{depth}$, camera intrinsic parameters $\bm{K}$, a ROI scaling factor  $\alpha_{\text{roi}}$
\STATE \textbf{Output}: 2D search region (ellipse)  $E$
\STATE \textbf{do}:
\STATE - $(^{C_{3D}}\bm{\mu}, ^{C_{3D}}\bm{\Sigma}) \leftarrow $ Project $(^M\bm{\mu}, ^M\bm{\Sigma})$ onto the 3D camera frame
\STATE - $(^{C_{2D}}\bm{\mu}, ^{C_{2D}}\bm{\Sigma}) \leftarrow $ Project $(^{C_{3D}}\bm{\mu}, ^{C_{3D}}\bm{\Sigma})$ onto the 2D camera frame using \eqref{eq:cam_projection} and \eqref{eq:cov_3d_to_2d}
\STATE -  $(\bm{\lambda_{u}}, \bm{\lambda_{v}}), (\bm{e_{u}}, \bm{e_{v}}) \leftarrow $ Extract eigenvalues and eigenvectors from $^{C_{3D}}\bm{\Sigma}$
\STATE - ($\bar{\lambda}_u, \bar{\lambda}_v$ $ \leftarrow $ Scale $ (\bm{\lambda_{u}}, \bm{\lambda_{v}}) $ by $ \alpha_{\text{roi}} $ to determine the axes length of the search ellipse (see  \eqref{eq:scaling})
\STATE - $ E \leftarrow $ Define the search ellipse centered at $^{C_{2D}}\bm{\mu}$ with axes $ \bm{\bar{\lambda}_{u}} \cdot \bm{e_{u}} $ and $  \bm{\bar{\lambda}_{v}} \cdot \bm{e_{v}} $
\IF{ellipse $ E $ falls within the image boundaries}
    \STATE - return E
\ENDIF
\STATE \textbf{end}
\end{algorithmic}
\end{algorithm}

\subsubsection{3D Position Calculation Using Search Ellipse in the Depth Image}
The final step is to search for the target inside the search region $E$ and calculate its 3D position measurement $^{KF}\bm{z}$ to be fed to the KF. Searching for the target basically means to find the group of depth pixels in $E$ that best represent that target. We first extract valid pixels within $E$. Valid pixels are defined such that they lie within 3-$\sigma_z$ of the depth uncertainty, which is extracted from $^{C_{3D}}\bm{\Sigma}$. Those valid pixels form a modified search ellipse $\bar{E}$. Then, we compute all possible contours inside $\bar{E}$, and select the one with the closest center to $^{C_{2D}} \bm{\mu}$. We denote this contour by $C_{\bar{E}}$ with center $\bm{\mu}_C = [u_{\bar{E}}, v_{\bar{E}}]^T$. This is to ensure that we avoid measurements that are too far from the latest KF estimate. Then, the average value $\bar{d}$ of the pixels inside $C_{\bar{E}}$, and its center $\bm{\mu}_C$ are used to calculate the 3D position measurement $^{C_{3D}}\bm{z}$ in the 3D camera frame using the camera matrix $\bm{K}$, equation \eqref{eq:contour_center_to_pos}.

\begin{equation}\label{eq:contour_center_to_pos}
    ^{C_{3D}}\bm{z} = \bm{K}^{-1} \cdot \begin{bmatrix} \bm{\mu}_C \\ 1 \end{bmatrix} \cdot \bar{d}
\end{equation}

Finally, the KF-guided 3D position measurement in the map frame $^{KF}\bm{z}$, to be fed to the KF , is calculated using the transformation matrix $ ^MT_{C_{3D}}$, equation \eqref{eq:kf_pos_2d_to_3d}.

\begin{equation}\label{eq:kf_pos_2d_to_3d}
    ^{KF}\bm{z} = ^MT_{C_{3D}} \cdot ^{C_{3D}}\bm{z}
\end{equation}

This final step is summarized in Algorithm \ref{alg:measurement_extraction}, which concludes this section.

\begin{algorithm}
\caption{Measurement Extraction from 2D Search Region} \label{alg:measurement_extraction}
\begin{algorithmic}
\STATE \textbf{Input}: $E$, $I_{depth}$, $\bm{K}$, $^MT_{C_{3D}}$ 
\STATE \textbf{Output}: 3D positions of detected target $^{KF}\bm{z}$
\STATE \textbf{do}:
\STATE - $\sigma_z \leftarrow$ Extract depth uncertainty from $^{C_{3D}}\bm{\Sigma}$
\STATE - $\bar{E} \leftarrow$ find valid pixels within $E$ that are within 3-$\sigma_z$
\STATE - $(C_{\bar{E}}, \mu_C) \leftarrow$ Find contours in $\bar{E}$, and extract the contour with closest center to $^{C_{2D}}\bm{\mu}$
\IF{found $(C_{\bar{E}}, \mu_C)$}
    \STATE - $^{C_{3D}}\bm{z} \leftarrow$ Calculate position measurement in 3D camera frame using \eqref{eq:contour_center_to_pos}
    \STATE - $^{KF}\bm{z} \leftarrow$ Calculate position in map frame using Equation (\ref{eq:kf_pos_2d_to_3d})
    \STATE - return $^{KF}\bm{z}$
\ELSE
    \STATE - Proceed without measurement for the current cycle
\ENDIF
\STATE \textbf{end}
\end{algorithmic}
\end{algorithm}


\section{Experiments and Results}\label{sec:results}

We validated the effectiveness of our proposed algorithms through a series of rigorous experiments conducted in a controlled simulation environment. Our simulation leverages several sophisticated tools to create a dynamic testing ground for our UAV tracking framework.

\subsection{Experimental Setup}
To ensure the reproducibility of our results, we detail our simulation setup as follows:

\begin{itemize}
\item \textbf{Simulation Platform}:

\textit{Gazebo Robotics Simulator}: Used for creating a virtual environment that hosts two quadcopter UAVs along with their sensory apparatus. Gazebo provides realistic physics and rendering to simulate UAV dynamics and sensor feedback accurately.
\textit{Version}: Specify the version of Gazebo used, as updates can introduce changes in physics calculations or feature sets.
\item \textbf{Autopilot Configuration}:

\textit{PX4 Autopilot}: Configured to control the trajectory of the target UAV. This open-source flight control software is critical for defining consistent and repeatable flight paths during simulations.
\textit{Firmware Version}: Include the firmware version of PX4 used, as firmware updates can affect UAV behavior.
\item \textbf{Software Integration}:

\textit{ROS 2}: Facilitates real-time data exchange and command sequences between the simulated devices and the control algorithms. It also ensures that the software developed in the simulation can be transferred seamlessly to real-world UAV operations.
\textit{ROS 2 Version}: Detail the specific release of ROS 2 used, to eliminate discrepancies caused by software updates.
\item \textbf{Hardware Specifications}:

\textit{Computer Specifications}: Experiments are run on a laptop equipped with an Nvidia RTX 3070 GPU and an Intel i7 processor, providing the necessary computational power to handle complex simulations without performance bottlenecks.
\textit{Operating System}: State the operating system and its version, as performance can vary across different platforms.
\item \textbf{Simulated Sensors}:

\textit{RGB and Depth Cameras}: Simulations include RGB and depth cameras modeled after the Intel Realsense D455. These cameras operate at 20Hz, outputting images at a resolution of 640x480 pixels.
\textit{Additional Sensors}: Simulated inertial measurement units (IMUs), GPS, and barometers are integrated into the UAVs to provide comprehensive data inputs necessary for precise position control via the PX4 autopilot.
\item \textbf{Environmental Conditions}:

Detail any environmental settings within Gazebo, such as lighting, weather conditions, or any physical obstacles introduced in the simulation environment, which can significantly impact sensor performance and UAV behavior.
\end{itemize}

\subsection{Scenarios}
Two scenarios were tested as follows where the target UAV was at a hover state and :
\begin{itemize}
    \item \textbf{Static Target}: The target UAV ascends to a height of 10 meters directly above the takeoff point and remains hovering in place. The observer UAV is stationed 15 meters 
away, with the target centrally within its FOV. This setup tests the framework’s ability to maintain accurate tracking with minimal target movement.

    \item {\textbf{Dynamic Target}}: The target UAV follows a circular trajectory with a 5-meter radius at a constant speed of 5 m/s and an altitude of \textbf{10 meters}. The test lasts for 2 minutes, to test the tracking system's capability to handle continuous target motion and directional changes.

\end{itemize}

For both scenarios, the observer UAV, functioning as a static aerial camera, attempts real-time state estimation of the target using the proposed algorithms (Section \ref{sec:algorithm}). The procedure for each simulation includes takeoff, positioning the observer to capture the target in its FOV, and executing the estimation framework. The Kalman filter estimates and associated errors are recorded for subsequent analysis.

\subsection{Results}

\subsubsection{Static Target Scenario}
\begin{figure} 
    \centering
    \includegraphics[width=\columnwidth]{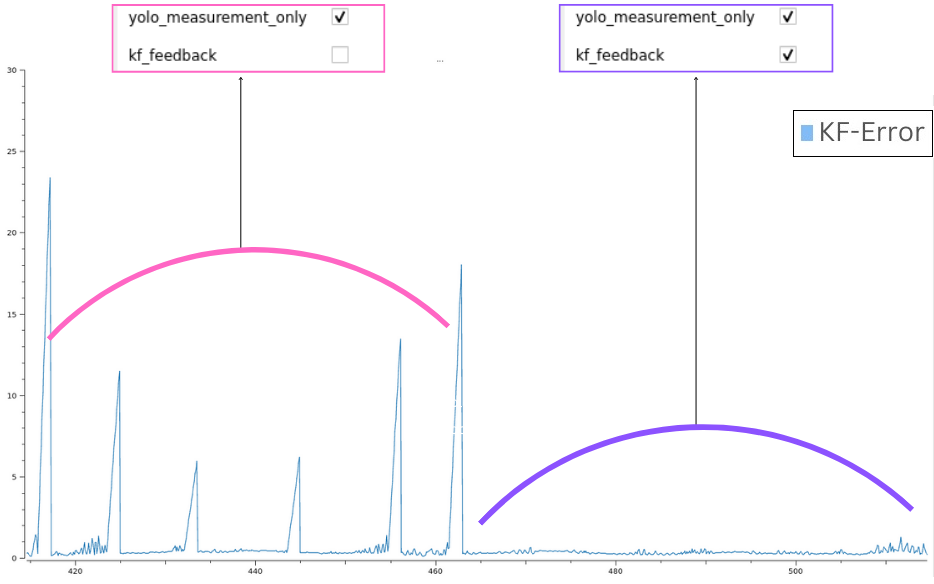}
    \caption{KF error vs. time for the static target scenario, with and without SMART-TRACK.}
    \label{fig:static}
\end{figure}

  Figure \ref{fig:static} presents the position error metrics, comparing the distance between true positions and Kalman Filter (KF) estimates when tracking a static target using both YOLOv8 and the SMART-TRACK framework. The graph illustrates significant peaks in estimation error (greater than 20 meters) when relying solely on YOLOv8 for target detection, which lacks feedback from the KF. These errors predominantly occur due to the limitations of YOLOv8 under conditions such as model inaccuracies, image blurring, or variable environmental effects.

However, the integration of KF feedback to guide the construction of search regions (Algorithm \ref{alg:search_region_extraction}) and to facilitate new measurements (Algorithm \ref{alg:measurement_extraction}) dramatically reduces these errors to less than 1 meter. This marked improvement underscores the efficacy of using depth data from pixels within the designated search region to estimate the target's position, even when YOLOv8 fails.

A critical insight from these observations is that the SMART-TRACK framework significantly compensates for the intermittent failures of the YOLOv8 detection system by leveraging Kalman Filter estimates to ensure continuity and accuracy in UAV tracking. The effectiveness of this approach highlights the potential for hybrid systems that combine traditional estimation techniques with advanced deep learning models to overcome the inherent weaknesses of each method when used in isolation.

\begin{figure} 
    \centering
    \includegraphics[width=\columnwidth]{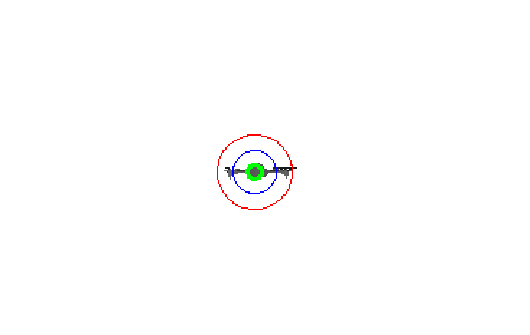}
    \caption{A search region with different scaling factors using \eqref{eq:scaling}, constructed using KF feedback measurements, and overlaid on a depth image with a target drone. $\alpha_{\text{roi}} = 1$ (smallest circle in green), $\alpha_{\text{roi}} = 3$ (middle circle in blue), $\alpha_{\text{roi}} = 5$ (outer circle in red)}
    \label{fig:roi_scaling}
\end{figure}

Moreover, it's important to note the need for scaling the search region $E$, as shown in equation \eqref{eq:scaling}, to ensure it encompasses a sufficient number of depth pixels that can robustly represent the target. Figure \ref{fig:roi_scaling} illustrates how different scaling factors $\alpha_{\text{roi}}$ affect the size of the search regions. An initial scaling factor of $\alpha_{\text{roi}}=1$, constructed using the eigenvalues and vectors of the projected 2D covariance matrix (\eqref{eq:cov_3d_to_2d}), provides a relatively narrow region that may be inadequate for effective target localization in the depth image. Our experiments suggest that a scaling factor of $\alpha_{\text{roi}}=5$ yields a more appropriately sized region, enhancing the reliability of subsequent target re-localization efforts within diverse operational environments.

This analytical approach not only confirms the robustness of the SMART-TRACK framework in static scenarios but also offers valuable methodological insights for optimizing tracking systems under similar static conditions. The application of a scaled search region, tailored by experimental insights, serves as a model for enhancing detection frameworks that can be tested further in more dynamic and unpredictable environments.

\subsubsection{Dynamic Target Scenario}

\begin{figure}
    \centering
    \includegraphics[width=\columnwidth]{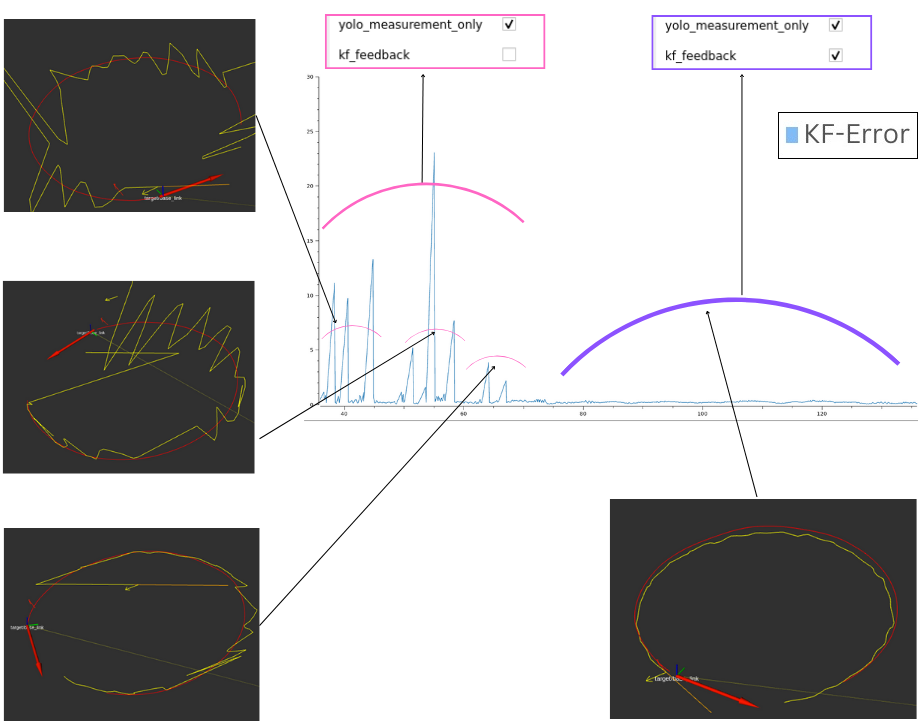}
    \caption{KF error vs. time for the dynamic target scenario, with and without KF$-$guided measurements.}
    \label{fig:Dynamic}
\end{figure}

Figure \ref{fig:Dynamic} illustrates the performance of the SMART-TRACK framework in tracking a target executing a circular trajectory. The path visualization clearly shows the effectiveness of the framework; the Kalman Filter (KF)-guided path (yellow) closely follows the target's actual path (red).

During the initial 15 seconds, the position error of the KF peaks significantly, exceeding 20 meters when relying solely on YOLOv8-based detection, as the target drone follows its circular route. This substantial error stems from intermittent measurements due to misdetections. However, once the KF feedback mechanism is activated to guide the search for new measurements, these gaps are efficiently augmented by KF-guided predictions, resulting in a dramatic reduction in tracking error to less than 1 meter. This substantial improvement underscores the KF's capability to maintain precise tracking of dynamically moving targets, even through complex maneuvers.

The initial challenges with YOLOv8 demonstrate the limitations of relying solely on AI-based detection for dynamic UAV tracking, particularly under motion-induced detection failures. The subsequent integration of KF feedback compensates for these deficiencies, stabilizing the tracking process by continuously updating the system with reliable position estimates when primary detections fail.

This adaptability is essential for real-world applications, suggesting that future enhancements should focus on refining the interaction between machine learning detection and model-based estimation techniques. Optimizing this integration can lead to more robust UAV tracking systems capable of operating under a broader range of environmental and operational conditions.

\subsubsection{Comparison with Prior Work}
To our knowledge, this study is the first to utilize Kalman Filter (KF) estimates for generating search regions to obtain new 3D position measurements of a UAV, particularly when primary measurements from the object detection module are unavailable. For a meaningful comparison, we benchmark our KF estimates against the most closely related works, specifically those presented in \cite{8756100} and \cite{vrba2020marker}, using the Root Mean Squared Error (RMSE) metric across various trajectories. It is crucial to acknowledge that the sensor specifications in our study may differ from those used in the referenced studies; nonetheless, this comparison aims to provide an understanding of the relative performance.

While alternatives such as the Extended Kalman Filter (EKF), Unscented Kalman Filter (UKF), particle filters, and LSTM-based methods offer advantages in addressing non-linear dynamics and trajectory prediction, they introduce significant computational overhead. Our approach prioritizes computational efficiency and real-time performance, making it more suitable for dynamic environments with high-speed targets where simpler models like constant velocity or acceleration are more practical. Additionally, our focus is on state estimation and measurement augmentation rather than trajectory prediction, which distinguishes our work from methods like LSTM. Accurate trajectory prediction for UAVs in 3D has been addressed in a different publication of ours currently under review.

We believe that our contribution of measurement augmentation using KF feedback effectively addresses the challenges of high-speed target tracking with constrained computational resources, and is well-suited for the applications discussed in this paper.

We calculated the average RMSE for our KF estimates compared to the true target positions over 100 iterations under both static and dynamic target scenarios. The results, presented in Table \ref{tab:trajectory_estimation}, reveal that the average RMSE in the static scenario is 0.31 m, and in the dynamic scenarios, it is 0.18 m for an infinity-like trajectory and 0.04 m for a circular trajectory. By contrast, the study in \cite{8756100}, which relies solely on depth images for UAV detection, reported an RMSE of 1.91 m for a target that is 2.13 m a way from the observer. Meanwhile, \cite{vrba2020marker}, which integrates modern AI-based detection methods including a Convolutional Neural Network (CNN), noted an RMSE of 3.76 m for a target that is 4.46 m away from the observer. These figures are significantly higher than those observed in our study across both static and dynamic scenarios, despite our farther target distance of $5.4m$. These results are tabulated in Table \ref{tab:trajectory_estimation-2}.

\begin{table}[ht]
  \centering
  \caption{KF estimation accuracy with and without KF-guided measurements}
  \label{tab:trajectory_estimation}
  \begin{tabular}{|c|c|c|c|c|}
    \hline
    Experiment & Method & RMSE & Distance \\
    \hline
    \multirow{2}{*}{Static} 
    & Yolo-Measurements-Only & 4.46 m & 4.2 m \\
    & KF-Feedback & 0.31 m & 4.2 m \\
    \hline
    \multirow{3}{*}{Dynamic} 
    & Yolo-Measurements-Only &  3.74 m & 5.4 m \\
    & KF-Feedback(inf) &  0.18 m & 5.4 m \\
    & KF-Feedback(cir) &  0.04 m & 5.4 m \\
    \hline
  \end{tabular}
\end{table}

\begin{table}[ht]
  \centering
  \caption{Comparison of Trajectory Estimation Methods}
  \label{tab:trajectory_estimation-2}
  \begin{tabular}{|c|c|c|}
    \hline
    Experiment & RMSE & Distance \\
    \hline
    \cite{8756100} depth-based &  1.9 m & 2.13 m  \\
    \cite{vrba2020marker} CNN & 3.76 m  & 4.46 m \\
    Ours-KF-Feedback(inf) & 0.18 m & 5.4 m \\
    Ours-KF-Feedback(cir) & 0.04 m & 5.4 m \\
    \hline
  \end{tabular}
\end{table}

\section{Conclusions}\label{sec:conclusions}

This study introduces a novel method for autonomous object tracking in unmanned aerial vehicles (UAVs), enhancing tracking accuracy in dynamic environments like security and drone-to-target interception. Our approach integrates YOLOv8 object detection with Kalman Filter-based estimation, addressing the limitations of traditional sensor-based tracking by maintaining high tracking stability even under conditions of intermittent target visibility.

SMART-TRACK utilizes high-frequency Kalman Filter estimates to guide the capture of new measurements when primary detection fails, substantially reducing tracking discontinuities. Tests in a ROS2-based simulation environment demonstrated our method's efficacy, achieving a root mean square error (RMSE) as low as 0.04 meters.

The integration of deep learning-based detection with traditional estimation techniques provides a resilient tracking system that adapts to rapid environmental changes and target movements. This hybrid approach significantly mitigates tracking errors and enhances system adaptability.

The system requires precise calibration between sensors and substantial computational resources, which may limit its use in resource-constrained scenarios. Performance in adverse weather or with severe sensor impairments remains to be fully explored.

Future research will aim to optimize the system’s computational demands for broader deployment and enhance its robustness against environmental challenges. Plans include incorporating additional sensor modalities, such as RADAR or LiDAR, to further improve tracking reliability across diverse conditions.

Overall, our research contributes to the field by significantly improving UAV tracking robustness and setting a foundation for more reliable autonomous UAV operations in complex settings.



\appendices

\section*{Acknowledgment}

The authors would like to thank Prince Sultan University for their support in providing the required equipment required for conducting the work described in this paper.

\bibliographystyle{IEEEtran}
\bibliography{references}

\end{document}